%% file: main.tex
\documentclass[letterpaper]{article} %
\usepackage{aaai2026}  %
\usepackage{times}  %
\usepackage{helvet}  %
\usepackage{courier}  %
\usepackage[hyphens]{url}  %
\usepackage{graphicx} %
\usepackage{natbib}  %
\usepackage{caption} %
\usepackage{pdfpages}
\usepackage{subcaption}
\usepackage{booktabs}
\input{preamble}

\usepackage{newfloat}
\usepackage{listings}
\DeclareCaptionStyle{ruled}{labelfont=normalfont,labelsep=colon,strut=off} %
\floatstyle{ruled}
\newfloat{listing}{tb}{lst}{}
\floatname{listing}{Listing}
\title{BOFA: Bridge-Layer Orthogonal Low-Rank Fusion for \\ CLIP-Based Class-Incremental Learning}
\author{
    Lan Li\textsuperscript{\rm 1,\rm 2},
    Tao Hu\textsuperscript{\rm 1,\rm 2},
    Da-Wei Zhou\textsuperscript{\rm 1,\rm 2}\thanks{Corresponding author.},
    Jia-Qi Yang\textsuperscript{\rm 1,\rm 2},
    Han-Jia Ye\textsuperscript{\rm 1,\rm 2},
    De-Chuan Zhan\textsuperscript{\rm 1,\rm 2}
}
\affiliations{
    \textsuperscript{\rm 1}School of Artificial Intelligence, Nanjing University, China\\
    \textsuperscript{\rm 2}National Key Laboratory for Novel Software Technology, Nanjing University, China

    \{lil, hut, zhoudw, yehj, yangjq, zhandc\}@lamda.nju.edu.cn
}

\usepackage{bibentry}

\begin{document}

\maketitle

\begin{abstract}
Class-Incremental Learning (CIL) aims to continually learn new classes without forgetting previously acquired knowledge. Vision-language models such as CLIP offer strong transferable representations via multi-modal supervision, making them a promising choice for CIL. However, applying CLIP to CIL poses two major challenges: (1) adapting to downstream tasks often requires additional learnable modules, increasing model complexity and susceptibility to forgetting; and (2) while multi-modal representations offer complementary strengths, existing methods have not fully exploited the synergy between visual and textual modalities. To address these issues, we propose BOFA (Bridge-layer Orthogonal Fusion for Adaptation), a novel framework for CIL. BOFA restricts adaptation to CLIP’s existing cross-modal bridge layer, keeping the core learning process parameter-free and avoiding any extra adaptation modules. To prevent forgetting within this layer, it leverages Orthogonal Low-Rank Fusion, a mechanism that constrains parameter updates to a low-rank ``safe subspace" that is mathematically constructed to be approximately orthogonal to the feature subspace of past tasks. This encourages stable knowledge accumulation and mitigates interference between new and previously learned classes. Furthermore, BOFA employs a cross-modal hybrid prototype that fuses stable textual prototypes with dynamic visual counterparts derived from our adapted bridge layer, resulting in a more robust and discriminative classifier. Extensive experiments on standard benchmarks demonstrate that BOFA achieves superior accuracy and efficiency compared to existing methods.
\end{abstract}

\begin{links}
    \link{Extended version}{https://arxiv.org/abs/2511.11421}
    \link{Code}{https://github.com/Lain810/BOFA}
\end{links}
\input{int}

\input{related}

\input{pre}
\input{method}

\input{exp}

\input{_con}

\section{Acknowledgements}

This work is partially supported by NSFC (62506160, 62376118), NSF of Jiangsu Province (BK 20251251, BK20243012), Fundamental Research Funds for the Central Universities (14380021), JSTJ-2025-147, Collaborative Innovation Center of Novel Software Technology and Industrialization.

\clearpage
\includepdf[pages=-]{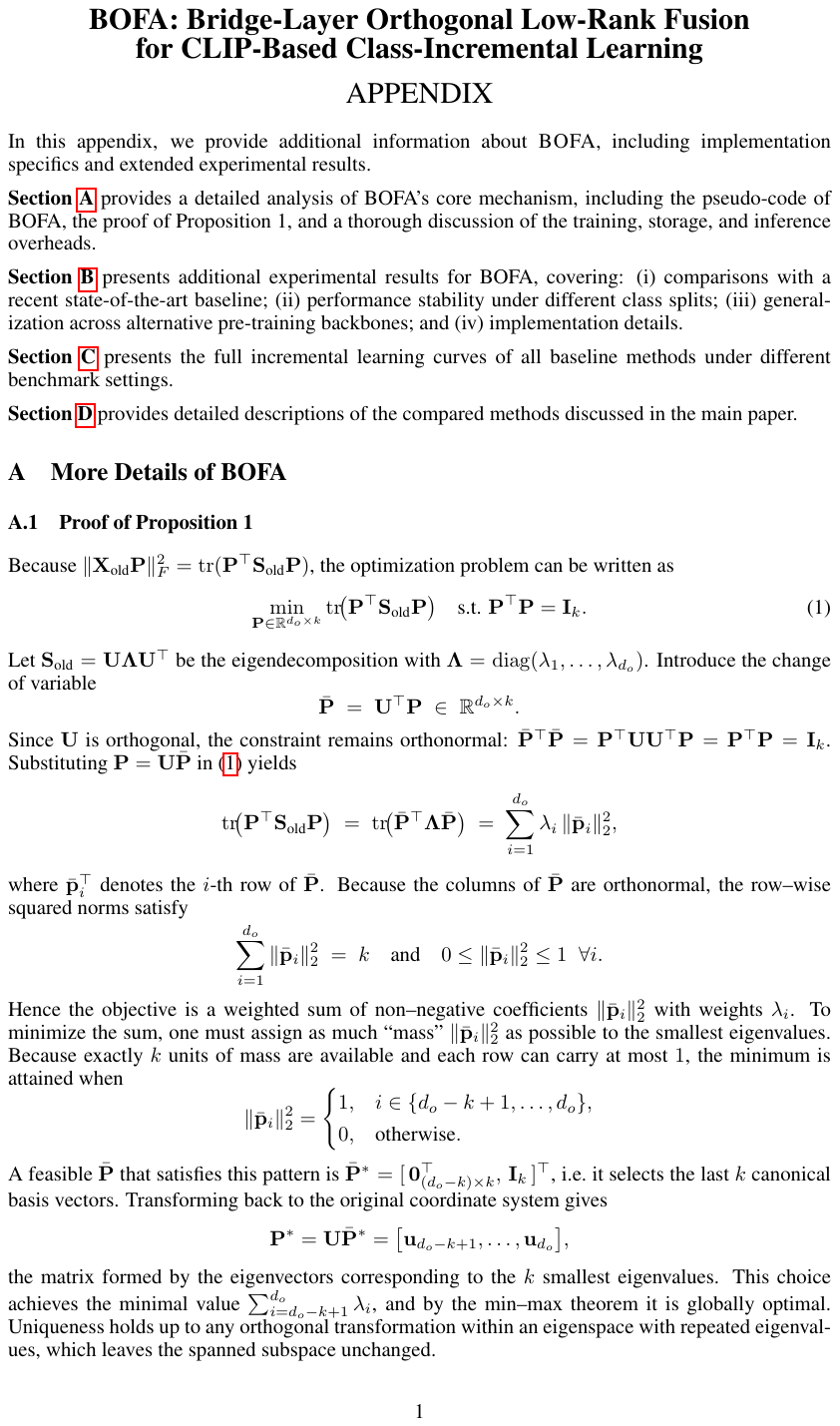}

\end{document}

%% file: preamble.tex
\usepackage{amsmath,amssymb} %
\usepackage{csquotes}
\usepackage{bm}
\usepackage{paralist,algorithm}
\usepackage[noend]{algpseudocode}
\usepackage{multirow}

\newcommand{\argmin}{\mathop{\arg\min}}

\newcommand{\name}{{\sc BOFA}}

\usepackage{accents}
\newlength{\dhatheight}

\usepackage{graphicx}

\usepackage{xcolor,colortbl}
\definecolor{Gray}{gray}{0.85}
\definecolor{Redo}{rgb}{0.95,0.69,0.51}
\definecolor{LightCyan}{rgb}{0.9, 0.9, 0.9}
\usepackage{afterpage}
\usepackage{pifont}%

\usepackage{csquotes}
\newtheorem{Proposition}{Proposition}

\usepackage{lipsum}
\definecolor{cvprblue}{rgb}{0.21,0.49,0.74}

%% file: int.tex
\section{Introduction}
In the real world, data is often encountered in a streaming fashion, where new classes arrive sequentially~\cite{aggarwal2018survey}. Class-Incremental Learning (CIL)~\cite{rebuffi2017icarl} addresses this challenge by enabling models to learn novel classes without forgetting previously acquired knowledge. However, continual learning systems are prone to catastrophic forgetting~\cite{french1999catastrophic,french1999modeling}, a phenomenon where learning new tasks interferes with performance on earlier ones. This problem is particularly pronounced when old data is unavailable due to privacy or storage constraints.

Recent advances in vision-language models (VLMs)~\cite{jia2021scaling,yang2023learning,li2024enhancing}, such as CLIP~\cite{radford2021learning}, have opened new possibilities for CIL. By aligning visual and textual modalities in a shared embedding space, VLMs provide rich, transferable representations that can generalize to unseen tasks. This makes them a compelling foundation for exemplar-free CIL. 

A prevailing strategy is to freeze the large backbones of CLIP to preserve general knowledge and only train a small set of additional modules, such as adapters or prompt layers, for new tasks~\cite{zhou2023learning,huang2024class}. While this approach maintains a stable high-level feature space, it introduces a new set of critical challenges. First, catastrophic forgetting persists within these trainable modules themselves. As adapters are sequentially fine-tuned on new tasks to perform cross-modal alignment, they overwrite knowledge crucial for old tasks, leading to a degradation in performance. Second, these additional parameters, however lightweight, result in additional inference cost, which poses a practical limitation for deployment in resource-constrained scenarios. Third, the training process for these modules is often suboptimal. They are typically trained via contrastive learning against textual prototypes derived from hand-crafted prompts. These generic prompts often lack the specificity needed to distinguish fine-grained visual concepts~\cite{wang2022learning,wang2023attriclip,zhou2023learning}, thereby limiting the model's discriminative capability.

To address these challenges, we propose BOFA (Bridge-layer Orthogonal Fusion for Adaptation), a novel and cohesive framework for efficient and continual adaptation of CLIP in CIL. Our approach first tackles the dual issues of inference overhead and catastrophic forgetting through a coupled strategy. The foundation of our method is to restrict all primary feature adaptation solely to CLIP's existing cross-modal bridge-layer, thereby bypassing the conventional need for external adaptation modules (e.g., adapters) and their associated costs.  This design, however, concentrates the risk of catastrophic forgetting within this single component. To counteract this, we introduce our primary technical contribution: Orthogonal Low-Rank Fusion. This mechanism constrains parameter updates to an ``Orthogonal Safe Subspace", a low-rank update subspace that is constructed to be approximately orthogonal to the feature subspace spanned by previously learned tasks. By projecting updates onto these non-interfering directions, BOFA ensures stable knowledge accumulation within the bridge-layer, effectively mitigating catastrophic forgetting without requiring data replay.
Building upon this stable adaptation, we introduce cross-modal hybrid prototypes to enhance classification. Our mechanism fuses static textual prototypes with dynamically refined visual prototypes, which are generated by our stably adapted bridge-layer. This synergy yields a robust and highly discriminative classifier, surpassing unimodal baselines.

In summary, our key contributions include: (1) We introduce Orthogonal Low-Rank Fusion, a novel adaptation method that mitigates catastrophic forgetting by constraining parameter updates within CLIP's bridge-layer to an orthogonal safe subspace, avoiding additional learnable parameters in the core adaptation mechanism and preserving the original model's architecture and inference path. (2) We design a cross-modal hybrid prototype that leverages our stable adaptation to fuse textual and visual prototypes, resulting in a more discriminative classifier. (3) The proposed BOFA achieves SOTA performance on multiple CIL benchmarks, demonstrating superior accuracy and efficiency.

%% file: related.tex
\section{Related Work}

\noindent\textbf{Class-Incremental Learning (CIL).}
CIL aims to enable models to incrementally learn new classes without forgetting previously acquired knowledge~\cite{masana2022class,de2021survey}. Traditional CIL approaches can be broadly categorized into several groups.
Distillation-based methods mitigate forgetting by aligning outputs between the current and previous models~\cite{hinton2015distilling}, through logit-level~\cite{rebuffi2017icarl,li2016learning}, feature-level~\cite{lu2022augmented,park2021class}, or group-level alignment~\cite{gao2022rdfcil,dong2021few}.
Replay-based methods explicitly preserve past knowledge by storing and revisiting samples from previous tasks~\cite{luo2023class}.
Regularization-based approaches estimate parameter importance and penalize updates to critical weights~\cite{aljundi2018memory,aljundi2019task}.
Bias rectification methods focus on correcting prediction or classifier bias accumulated during incremental training~\cite{shi2022mimicking}.
Model expansion methods dynamically increase model capacity by expanding neurons~\cite{yoon2018lifelong,xu2018reinforced}, backbones~\cite{zhou2022model,wang2023beef,zheng2025task}, or lightweight components~\cite{douillard2022dytox} to accommodate new knowledge.

\noindent\textbf{Pre-Trained Model-Based CIL.}
Leveraging pre-trained models has emerged as a promising direction for CIL~\cite{zhou2025duct,qi2025adaptive,sun2024mos,zhou2025external}, as they offer generalizable representations that can accelerate learning and improve performance. A prevalent strategy is to freeze the pre-trained backbone and introduce lightweight, learnable modules such as prompts~\cite{wang2022dualprompt,wang2022learning,smith2023coda,wang2022s,zhou2022learning} and adapters~\cite{chen2022adaptformer,yu2024boosting,wang2025integrating}.
L2P~\cite{wang2022learning} and DualPrompt~\cite{wang2022dualprompt} utilize learnable prompt pools with selection mechanisms tailored for pre-trained vision transformers~\cite{jia2022visual}. Further extensions explore more sophisticated prompt composition using attention~\cite{smith2023coda} or generative networks~\cite{jung2023generating}.
Another line of work builds classifiers directly on top of pre-trained embeddings, matching class prototypes to feature representations~\cite{zhou2023revisiting,mcdonnell2023ranpac,snell2017prototypical,li2025addressing}.
For multi-modal settings with pre-trained CLIP, several approaches design cross-modal prompt tuning schemes to enhance alignment~\cite{wang2023attriclip,wang2022s}.
MOE-Adapter~\cite{yu2024boosting} introduces mixture-of-experts mechanisms~\cite{masoudnia2014mixture} for adaptive module selection, while PROOF~\cite{zhou2023learning} extends CLIP's representational capacity with task-specific projection heads.
RAPF~\cite{huang2024class} further decomposes parameter updates into modular components, enabling adaptive adapter fusion to balance plasticity and stability.

%% file: pre.tex
\section{Preliminaries}
This section introduces the exemplar-free CIL setup and outlines how CLIP can be adapted to incremental learning.
 \subsection{Class-Incremental Learning}
 Class-incremental learning (CIL) aims to incrementally construct a classifier that is capable of recognizing all classes encountered over a sequence of tasks in a data stream setting~\cite{rebuffi2017icarl}. We denote the sequence of training sets as $\{\mathcal{D}^1, \mathcal{D}^2, \ldots, \mathcal{D}^T\}$, where the dataset for task $t$ is given by $\mathcal{D}^t = \{(\mathbf{x}_i, y_i)\}_{i=1}^{n_t}$.  Given this setup, each dataset consists of input-label pairs $(\mathbf{x}_i, y_i)$, where each input $\mathbf{x}_i \in \mathbb{R}^D$ is associated with a label $y_i$ drawn from the task-specific label set $Y_t$, with $Y_t \cap Y_{t'} = \varnothing$ for $t \neq t'$.
 
This work focuses on the \textbf{exemplar-free} CIL setting, where no historical instances can be stored for rehearsal~\cite{zhu2021prototype, wang2022learning, wang2022dualprompt}. Consequently, during the $t$-th incremental step, the model has access only to the current task data $\mathcal{D}^t$. The objective is to train a classifier $f$ over the union of all seen label sets $\mathcal{Y}_t = \bigcup_{t=1}^{T} Y_t$ that minimizes the expected misclassification risk:
\begin{equation*}
f^* = \argmin_{f \in \mathcal{H}} \mathbb{E}_{(\mathbf{x}, y) \sim \bigcup_{t=1}^{T} \mathcal{D}^t} \bigl[\mathbb{I}(y \neq f(\mathbf{x}))\bigr],
\end{equation*}
where $\mathcal{H}$ is the hypothesis space, $\mathbb{I}(\cdot)$ is the indicator function, and $\mathcal{D}^t$ denotes the data distribution of task $t$.  
\subsection{CLIP-Based CIL}
We build upon CLIP~\cite{radford2021learning}, a VLM pretrained on image–text pairs, as the foundation for our CIL framework, following prior work~\cite{zhou2023learning,yu2024boosting,huang2024class}.

\noindent{\bf CLIP architecture.} 
CLIP employs two modality-specific encoders to project images and texts into a shared $d$-dimensional embedding space: a text encoder and an image encoder. The image encoder is typically decomposed as $g_i = g_2 \circ g_1$, where $g_1$ is a visual backbone (e.g., ViT) that extracts raw visual features $\mathbf{x}_o \in \mathbb{R}^{d_o}$. The subsequent layer, $g_2$, is a linear projection that maps these features into the joint embedding space. We refer to $g_2$, parameterized by its weight matrix $\mathbf{W} \in \mathbb{R}^{d_o \times d}$, as the \textbf{cross-modal bridge-layer}. The final image embedding is thus computed as
\begin{equation*}
\mathbf{x}_o = g_1(\mathbf{x}), \quad 
\mathbf{z}_i = g_2(\mathbf{x}_o) =  \mathbf{x}_o\mathbf{W},
\end{equation*}
\noindent{\bf Zero-Shot Classification with Textual Prototypes.} A key capability of CLIP is its zero-shot classification performance. For a given set of classes, a classifier can be constructed on-the-fly from their names. Each class name is embedded into a textual prompt, such as ``a photo of a \texttt{[CLASS]}," and then encoded by $g_t$ to form a \textbf{textual prototype}. An input image $\mathbf{x}$ is classified by computing the cosine similarity between its visual embedding $\mathbf{z}_i$ and each textual prototype $\mathbf{z}_t^c$:
\begin{equation}
P(y=c \mid \mathbf{x}) = \frac{\exp(\cos(\mathbf{z}_i, \mathbf{z}_t^c) / \tau)}{\sum_{c'} \exp(\cos(\mathbf{z}_i, \mathbf{z}_t^{c'}) / \tau)}, \label{eq:clip_classifier}
\end{equation}
where $\tau$ is a temperature parameter.

\noindent{\bf Adapting CLIP for Incremental Learning.}
While the zero-shot classifier provides a strong baseline, the model's performance can be further improved by fine-tuning on downstream training data, typically using the cross-entropy loss on the probabilities in Eq.~\eqref{eq:clip_classifier}. In the context of CIL, a prevalent adaptation strategy is to freeze the large, pre-trained backbones ($g_i$ and $g_t$)	to preserve their general knowledge. To learn task-specific information, a lightweight, trainable module, such as an adapter, is introduced after the image encoder. This module is then sequentially updated as new tasks arrive.

\noindent{\bf Discussion.}
Although CLIP provides a promising foundation for CIL, its adaptation faces several key challenges. First, the dominant approach of adding an external adapter, while protecting the model's backbone, merely relocates the problem of catastrophic forgetting to the adapter itself. As this compact module is sequentially updated, it still struggles to retain knowledge of previous tasks, leading to significant performance degradation. Moreover, this approach introduces extra parameters and inference latency, compromising model efficiency. Second, the reliance on textual prototypes makes the classifier's quality highly sensitive to prompt engineering and often results in suboptimal alignment with data-specific visual features. Therefore, an effective CLIP-based CIL approach should mitigate forgetting without resorting to external adaptation modules, while also robustly leveraging cross-modal information.

%% file: method.tex
\section{BOFA: Bridge-layer Orthogonal Fusion for Adaptation}
To tackle the challenges of catastrophic forgetting and suboptimal modality alignment discussed above, we design our \name~around three key components: (1) lightweight fine-tuning of CLIP’s bridge-layer, (2) orthogonal low-rank fusion to preserve previous knowledge, and (3) cross-modal hybrid prototypes to enhance classification performance.
\subsection{Fine-tuning the Cross-Modal Bridge-Layer}
Most prior CLIP-based CIL methods introduce external trainable modules (e.g., adapters or prompt layers) to bridge the gap between new visual and textual concepts~\cite{zhou2023learning,huang2024class}. In contrast, we propose a streamlined approach where all {core feature adaptation} is performed exclusively on CLIP's existing parameters. Specifically, we fine-tune only the projection layer $g_2$, which serves as the cross-modal bridge-layer that maps the high-dimensional visual features $\mathbf{x}_o = g_1(\mathbf{x})$ extracted by the frozen visual backbone $g_1$ into the shared embedding space. By restricting updates to this single, pre-existing layer while keeping both the visual backbone $g_1$ and the text encoder $g_t$ frozen, we \textbf{preserve the original CLIP architecture and avoid inserting new computational layers into the main inference path}.
This adaptation leverages the semantic richness and generality of the high-dimensional visual feature space $\mathbb{R}^{d_o}$, prior to projection, enabling flexible and efficient learning of new tasks. Consequently, the forward pass for generating the primary visual embeddings {incurs no additional cost} compared to the base CLIP model, making our {core adaptation mechanism} exceptionally efficient.
\subsection{Orthogonal Low-Rank Fusion}
While fine-tuning the cross-modal bridge-layer enables efficient adaptation to new tasks, preserving past knowledge remains a critical challenge in CIL. Naive sequential fine-tuning disrupts previously learned image-text alignments, causing catastrophic forgetting. To address this issue, we propose orthogonal low-rank fusion, a framework that constrains fine-tuning updates to a principled, low-dimensional subspace to minimize interference with past tasks. By leveraging the approximate null space of previous task features, our method effectively preserves past knowledge while adapting to new tasks. This constraint is then efficiently implemented using a modified Low-Rank Adaptation (LoRA)~\cite{hu2022lora} scheme.

\noindent{\textbf{Forgetting Analysis.}} Let $\mathbf{W}_{\text{old}} = \mathbf{W}_{0} + \Delta \mathbf{W}_{\text{old}}$ be the weights of the bridge-layer after training on a sequence of past tasks, where $\mathbf{W}_{0}$ is the pretrained weight and $\Delta \mathbf{W}_{\text{old}}$ is the fused parameter update of past tasks. To adapt to a new task, fine-tuning produces a parameter update $\Delta \mathbf{W}_{\text{new}}$ of the new task. After adaptation, the fused weight matrix becomes $\mathbf{W}_{\text{new}} = \mathbf{W}_{\text{old}} + \Delta \mathbf{W}_{\text{new}}$.  For the feature matrix $\mathbf{X}_{\text{old}}$ collected from previous tasks (each row representing the visual feature $\mathbf{x}_{\text{o}}$ of a previous sample), the projected embeddings are perturbed from $\mathbf{X}_\text{old} \mathbf{W}_{\text{old}} $ to: 
\begin{equation*} 
\mathbf{X}_{\text{old}} \mathbf{W}_{\text{new}}  = \mathbf{X}_{\text{old}}\mathbf{W}_{0} + \mathbf{X}_{\text{old}}\Delta\mathbf{W}_{\text{old}} + \underbrace{ \mathbf{X}_{\text{old}}\Delta\mathbf{W}_{\text{new}}}_{\text{Interference Term}}. 
\end{equation*} 
The interference term is the primary source of forgetting. If the cumulative effect of this interference across all past features is significant, the representations of previous tasks are substantially degraded. To mitigate this, the core idea is to constrain the update $\Delta \mathbf{W}_{\text{new}}$ such that its effect on the feature space of previous tasks, represented by the feature matrix $\mathbf{x}_{o,\text{old}}$, is minimized. Ideally, we seek to satisfy: 
\begin{equation} \mathbf{X}_{\text{old}} \Delta \mathbf{W}_{\text{new}} \approx \mathbf{0}. \label{eq:null_condition} 
\end{equation} 
This condition implies that each column of the update matrix $\Delta \mathbf{W}_{\text{new}}$ should belong to the \textbf{null space} of $\mathbf{X}_{\text{old}}$, thereby preserving knowledge of past tasks while learning new ones.

\paragraph{Constructing an Orthogonal Safe Subspace.} In practice, the high-dimensional feature matrix $\mathbf{X}_{\text{old}}$, comprising diverse data from multiple tasks, is typically full-rank. Consequently, its exact null space is trivial (containing only the zero vector), which renders the strict constraint in Eq.~\eqref{eq:null_condition} impractical. To address the challenge of catastrophic forgetting, we relax the notion of an exact ``null space" into an approximate space. Specifically, we define an approximate null space as any subspace into which the projection of previous task features has low magnitude. Constraining weight updates to lie within such a subspace effectively minimizes interference with representations from previous tasks.  

Let $\mathbf{P} \in \mathbb{R}^{d_o \times k}$ represent an orthonormal basis matrix such that $\mathbf{P}^\top \mathbf{P} = \mathbf{I}_k$. We measure the interference induced by this basis on the old features using the Frobenius norm:
\[
\mathcal{I}(\mathbf{P}) = \bigl\| \mathbf{X}_{\text{old}} \mathbf{P} \bigr\|_F^2,
\]
and define the optimal $k$-dimensional approximate null space as:
\begin{equation}
\label{eq:approx_null_space}
\mathbf{P}^* = \arg\min_{\mathbf{P}^\top \mathbf{P} = \mathbf{I}_k} \bigl\| \mathbf{X}_{\text{old}} \mathbf{P} \bigr\|_F^2.
\end{equation}

 \begin{Proposition} \label{prop:evectors} 
 The cumulative scatter matrix of past features is defined as \(\mathbf{S}_{\text{old}} = \mathbf{X}_{\text{old}}^\top \mathbf{X}_{\text{old}}\), which can be decomposed as: 
 \[ \mathbf{S}_{\text{old}} = \mathbf{U} \operatorname{diag}(\lambda_1, \dots, \lambda_{d_o}) \mathbf{U}^\top, \quad \lambda_1 \geq \cdots \geq \lambda_{d_o}.\] 
 The optimal solution to Eq.~\eqref{eq:approx_null_space} is given by: 
 \[ \mathbf{P}^* = \bigl[\mathbf{u}_{d_o-k+1}, \dots, \mathbf{u}_{d_o}\bigr], \] 
which is the subspace spanned by the eigenvectors of \(\mathbf{S}_{\text{old}}\) associated with its \(k\) smallest eigenvalues.
 \end{Proposition}

We term the subspace spanned by $\mathbf{P}^*$ the Orthogonal Safe Subspace (OSS). It represents the $k$-dimensional subspace that minimizes the projection of past features, thus identifying the directions of least interference. To mitigate catastrophic forgetting, we constrain the parameter update $\Delta{\mathbf{W}_\text{new}}$  to lie within this subspace. We subsequently employ a modified LoRA scheme to implement this constraint efficiently, as detailed below. A full proof of Proposition~\ref{prop:evectors} is provided in extended version.

It is worth noting that in CIL scenarios, maintaining the OSS does not require storing all previous task features. The cumulative scatter matrix is incrementally updated as:
\[
\mathbf{S}_\text{new} = \mathbf{S}_{\text{old}}+\mathbf{X}_\text{new}^\top \mathbf{X}_\text{new}.
\]
The updated OSS is then efficiently computed from the $k$ smallest eigenvectors of $\mathbf{S}_{\text{new}}$, thus avoiding any dependency on a data replay buffer.

\begin{figure}[t]
	\centering
	\includegraphics[width=\linewidth]{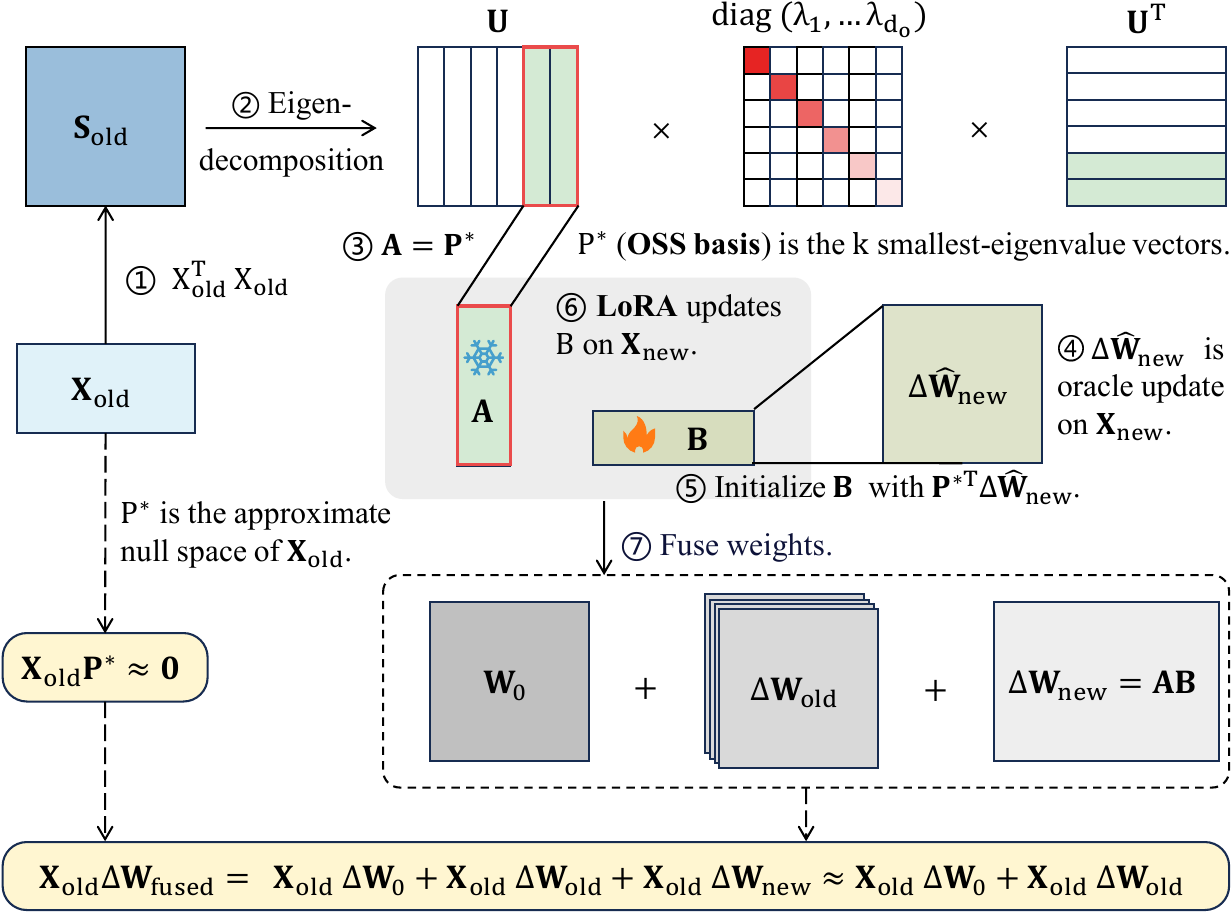}
\captionsetup{skip=8pt}
    \caption{{Overview of Orthogonal Low-Rank Fusion, where an OSS $\mathbf{P}^*$ is constructed from past task features to constrain the low-rank update for a new task, thereby minimizing interference with prior knowledge. }}
	\label{olrf}
\end{figure}
\paragraph{LoRA in the Orthogonal Safe Subspace.} To implement the update constraint efficiently, we adapt the LoRA framework. Standard LoRA approximates the weight update as $\Delta\mathbf{W} = \mathbf{A}\mathbf{B}$, where  $\mathbf{A} \in \mathbb{R}^{d_o \times k}$ and $\mathbf{B} \in \mathbb{R}^{k \times d}$ are trainable low-rank matrices. Our goal is to enforce that the column space of $\Delta\mathbf{W}$ lies within the OSS. Since the rows of $\Delta\mathbf{W}$ are linear combinations of the rows of $\mathbf{A}$, we fix the low-rank matrix $\mathbf{A}$ to be the OSS basis, i.e., we set $\mathbf{A} = \mathbf{P}^*$, where $\mathbf{P}^*$ is the orthonormal basis spanning the approximate null space as defined above. With this parameterization, the weight update is given by $\Delta\mathbf{W} = \mathbf{P}^*\mathbf{B}$, where only $\mathbf{B}$ remains trainable.

However, when $\mathbf{A}$ is frozen to the orthogonal safe subspace basis, simply initializing $\mathbf{B}$ to zero can lead to optimization difficulties, as the initial optimization direction may not align with the new task's objectives within the OSS. To address this, we use a data-driven initialization for $\mathbf{B}$. We first obtain a temporary ``oracle'' update $\Delta\tilde{\mathbf{W}}_{\text{new}}$ by briefly fine-tuning the entire bridge-layer on the new task data. Our goal is to initialize $\mathbf{B}$ by minimizing the discrepancy $\| \Delta\tilde{\mathbf{W}}_{\text{new}} - \mathbf{A}\mathbf{B} \|_F^2$ with $\mathbf{A}$ frozen. When $\mathbf{A}$ is set to the basis $\mathbf{P}^*$, this yields the closed-form solution
\begin{equation*}
\mathbf{B}_0 = \mathbf{P}^{*\top}\Delta\tilde{\mathbf{W}}_{\text{new}}.
\end{equation*}
This initialization provides a safe and task-adaptive update that follows the oracle direction within the OSS, which accelerates convergence and improves optimization stability.

Figure~\ref{olrf} provides an overview of the complete orthogonal low-rank fusion pipeline, including the construction of the OSS and its LoRA-based implementation. After the current task is learned, the bridge-layer parameters are updated as:
\begin{equation*} 
\mathbf{W}_{\text{fused}}  = \mathbf{W}_{0} + \Delta\mathbf{W}_{\text{old}} + \Delta\mathbf{W}_{\text{new}}. 
\end{equation*} 
\subsection{Learning with Cross-Modal Hybrid Prototypes}
As established in our preliminaries, relying solely on textual prototypes can limit classification performance. To address this, we introduce a cross-modal hybrid classifier that fuses general semantic knowledge from text with data-driven visual characteristics. 

\noindent{\bf{Static Hybrid Prototype Construction.}}
For each class $c$, we first construct a static hybrid prototype $\mathbf{p}_c$ by linearly interpolating between its textual prototype $\mathbf{z}_t^c$ and its visual prototype $\mathbf{z}_i^c$:
\begin{equation}
  \mathbf{p}_c = (1-\lambda)\,\mathbf{z}_t^c + \lambda\,\mathbf{z}_i^c.
\label{eq:fused_prototype}
\end{equation}
The textual prototype $\mathbf{z}_t^c$ is the pre-computed embedding from the frozen text encoder $g_t$. The visual prototype $\mathbf{z}_i^c$ is obtained by applying the bridge-layer $g_2$ to the mean high-dimensional feature $\bar{\mathbf{x}}_o^c$ of class $c$. The balancing coefficient $\lambda$ is selected by grid search using the training data of the first task and is then kept fixed for all subsequent tasks.

\noindent{\bf{Dynamic Prototype Refinement.}}
The continuous adaptation of the bridge-layer $g_2$ causes representation drift, which makes previously computed visual prototypes increasingly outdated. To address this, we introduce a dynamic refinement process~\cite{li2024twice}. During training on a new task, we use an Exponential Moving Average (EMA) to continuously update the visual prototypes of all seen classes. This provides a stable, up-to-date set of prototype embeddings for the classifier, thereby stabilizing training. After all tasks have been learned, we perform a final one-time refinement by regenerating all visual prototypes from their stored mean high-dimensional features using the final bridge-layer $\mathbf{W}_{\text{fused}}$. This ensures the final classifier is aligned with the final state of the model used for evaluation.

\noindent{\bf{Hierarchical Inference.}}
As the number of classes $|\mathcal{Y}|$ grows, the final classifier becomes more prone to inter-class confusion. To mitigate this, we employ a hierarchical classification strategy. While our core adaptation mechanism is parameter-free, this strategy introduces a small set of ancillary parameters primarily to enhance accuracy. It first uses a bank of lightweight, task-specific linear classifiers on features $\mathbf{x}_o$ to form a candidate subset, composed of the top-1 prediction from each task. The primary classifier then performs its final discrimination exclusively on this pruned subset. This structured approach reduces ambiguity by narrowing the search space, leading to improved performance.

\subsection{Discussion of BOFA}

In summary, BOFA's methodology rests on three synergistic components. 
First, its core feature adaptation is confined to CLIP's existing bridge-layer, avoiding external adaptation modules. Second, to protect past knowledge, our proposed orthogonal low-rank fusion constrains this layer's updates to an Orthogonal Safe Subspace that minimizes interference. Finally, we enhance classification accuracy using {data-driven hybrid prototypes}, which are deployed within a {hierarchical strategy that employs auxiliary classifiers} to mitigate inter-class confusion.This integrated design is highly parameter-efficient. Its primary storage overhead stems from the cumulative scatter matrix ($\mathbb{R}^{d_o \times d_o}$), the mean high-dimensional feature for each class ($|\mathcal{Y}| \times d_o$), and the lightweight auxiliary classifiers. 
This is a significant improvement over methods requiring data replay like PROOF~\cite{zhou2023learning} or large per-class covariance matrices (approx. $|\mathcal{Y}| \cdot d^2$) like RAPF~\cite{huang2024class}. Crucially, by keeping the {core inference path of CLIP unaltered}, BOFA remains a practical and highly scalable solution for CIL. A detailed cost analysis and pseudocode can be found in the extended version.

%% file: exp.tex
\section{Experiments}
\input{Tabs/tab1}

\subsection{Implementation Details}

\noindent {\bf Dataset}: Following prior works~\cite{zhou2023learning,zhou2022learning,wang2022learning}, we evaluate performance on nine benchmark datasets that exhibit significant domain shift from CLIP’s pre-training data. These include: \textit{CIFAR100}~\cite{krizhevsky2009learning}, \textit{CUB200}~\cite{WahCUB2002011}, \textit{ObjectNet}~\cite{barbu2019objectnet}, \textit{ImageNet-R}~\cite{hendrycks2021many}, \textit{FGVCAircraft}~\cite{maji2013fine}, \textit{StanfordCars}~\cite{krause20133d}, \textit{Food101}~\cite{bossard2014food}, \textit{SUN397}~\cite{xiao2010sun}, and \textit{UCF101}~\cite{soomro2012ucf101}.
To facilitate class-incremental splits, we follow the sampling strategy in~\citet{zhou2023learning}: selecting 100 classes from CIFAR100, Aircraft, Cars, Food101, and UCF101; 200 classes from CUB200, ObjectNet, and ImageNet-R; and 300 classes from SUN397.
Detailed dataset statistics and splits are provided in the extended version.

\noindent {\bf Dataset split:} Following the convention in~\cite{rebuffi2017icarl,wang2022learning}, we adopt the B-$m$ Inc-$n$ protocol~\cite{zhou2023learning,zhou2025external} to simulate CIL. Here, $m$ denotes the number of classes introduced in the initial base session, and $n$ represents the number of new classes added at each subsequent incremental stage.
To ensure reproducibility and consistency, we randomly shuffle the class order using a fixed seed (1993), as in~\cite{rebuffi2017icarl}, and apply the same ordering across all methods.

\noindent {\bf Comparison methods:} We compare our approach with several state-of-the-art CIL methods that leverage pre-trained models, including L2P~\cite{wang2022learning}, DualPrompt~\cite{wang2022dualprompt}, CODA-Prompt~\cite{smith2023coda}, and SimpleCIL~\cite{zhou2023revisiting}. In addition, we evaluate against recent CLIP-based CIL approaches such as CoOp~\cite{zhou2022learning}, PROOF~\cite{zhou2023learning}, and RAPF~\cite{huang2024class}. A naive baseline, denoted as Finetune, directly fine-tunes CLIP on the incremental tasks. All methods are initialized from the same CLIP model to ensure a fair comparison.

\noindent {\bf Training details:} All experiments are conducted using PyTorch~\cite{paszke2019pytorch} on an NVIDIA RTX 4090 GPU. Following prior work~\cite{zhou2023learning,huang2024class}, we adopt the CLIP ViT-B/16 model as the visual backbone across all methods. For vision-only methods that cannot utilize text prompts (e.g., L2P, DualPrompt, CODA-Prompt), we initialize them using CLIP’s visual encoder.
Unless otherwise stated, we report results using CLIP pre-trained on LAION-400M~\cite{ilharco_gabriel_2021_5143773}. Our method is trained via a 15-epoch fine-tuning stage for initialization, then a 5-epoch stage for LoRA training within the OSS. We use SGD with a batch size of 128 and a cosine-annealed learning rate starting from 0.05. The rank $k$ is set to 64. More experimental details are provided in the extended version. 
\input{Figs/curv}

\noindent {\bf Evaluation metric:} Following established protocols~\cite{rebuffi2017icarl,zhou2023learning}, we evaluate the top-1 accuracy after each incremental stage $b$, denoted as $\mathcal{A}_b$. We report the final accuracy after the last stage, $\mathcal{A}_B$, as well as the average accuracy across all stages, $\bar{\mathcal{A}} = \frac{1}{B}\sum_{b=1}^{B}\mathcal{A}_b$.
\subsection{Benchmark Comparison}
We first evaluate \name~against a range of state-of-the-art methods on standard CIL benchmarks. Results are presented in Table~\ref{tab:benchmark} and visualized in Figure~\ref{fig:curve}. \name~consistently outperforms existing approaches across all datasets, demonstrating its robust generalization and continual learning capabilities. Among all methods, the naive fine-tuning baseline performs the worst, suggesting that without proper regularization, the model completely forgets previously learned class representations. Visual prompt-based methods such as L2P, DualPrompt, and CODA-Prompt show limited performance, likely due to their inability to incorporate semantic information from the textual modality. In contrast, CoOp, a textual prompt tuning approach, suffers from substantial degradation in performance, which we attribute to severe forgetting of learned prompts over time. Even compared to recent CLIP-based approaches like RAPF, \name~achieves significant gains, demonstrating superior robustness against catastrophic forgetting while simultaneously preserving the inherent benefits of cross-modal representations.

In addition to exemplar-free methods, we also compare against representative exemplar-based CIL approaches, include iCaRL~\cite{rebuffi2017icarl}, MEMO~\cite{zhou2022model} and PROOF~\cite{zhou2023learning}. These results can be found in the extended version.
\subsection{Further Analysis}
\noindent \textbf{Ablation Study of Orthogonal Low-Rank Fusion:}
To validate our orthogonal low-rank fusion, we benchmark several methods applied specifically to the cross-modal bridge-layer: sequential fine-tuning, standard LoRA, and an adapted version of RAPF. For RAPF, a state-of-the-art fusion method, we applied its core logic and removed its covariance-based sampling to ensure a fair comparison. As shown in Figure~\ref{fig:ablation}, BOFA significantly outperforms all baselines. Notably, our results reveal that naive fine-tuning surpasses standard LoRA. This suggests that for adapting the bridge-layer, a simple low-rank constraint is overly restrictive for learning, yet insufficient for preventing forgetting. Our approach is engineered to resolve this dilemma. It harnesses the adaptive power of fine-tuning to find a strong initial update, then projects this update into the Orthogonal Safe Subspace, nullifying its interference with past knowledge while preserving its task-adaptive direction. This synergy of targeted plasticity and principled stability is the key to BOFA's superior performance.

\input{Figs/fig1}

\input{Tabs/tab3}

\noindent \textbf{Effectiveness of Cross-Modal Hybrid Prototypes:}
Table~\ref{tab:proto} validates our strategy against two single-modality baselines. These variants use purely textual (Textual) or visual (Visual) prototypes while retaining other BOFA components. Results reveal single-modality limitations: Textual prototypes perform better on ImageNet-R, whereas Visual ones excel on SUN, indicating dataset-specific biases. In contrast, our hybrid model consistently outperforms both variants. This demonstrates that fusing textual semantics with visual characteristics creates a more robust strategy, confirming that cross-modal fusion is vital for \name's success.
\input{Figs/feat}

\noindent\textbf{Visualizations}:
We employ t-SNE~\cite{van2008visualizing} to visualize cross-modal features learned by \name~on CIFAR100 B0 Inc5 in Figure~\ref{fig:tsne}. We compare feature distributions with and without Orthogonal Low-Rank Fusion on the second task, plotting features ($\bullet$) from 5 old and 5 new test classes, along with their corresponding prototypes ($\bigstar$).
Without fusion, new class features are well-clustered, but old class features remain entangled and difficult to distinguish. After fusion, features from both old and new classes show clearer separation and better alignment with their prototypes, indicating improved discriminability and prototype consistency.

%% file: Tabs/tab1.tex
\begin{table*}[t]

	\centering
	\resizebox{1\textwidth}{!}{%
		\begin{tabular}{@{}lccccccccccccccc}
			\toprule
			\multicolumn{1}{c}{\multirow{3}{*}{Method}}
			&
			\multicolumn{4}{c}{Aircraft }   & 
			\multicolumn{4}{c}{CIFAR100 }	&	
			\multicolumn{4}{c}{Cars }   
			\\ 
			& 
			\multicolumn{2}{c}{B0 Inc10}   & 
			\multicolumn{2}{c}{B50 Inc10}	&		
			\multicolumn{2}{c}{B0 Inc10}   & 
			\multicolumn{2}{c}{B50 Inc10}	& 
			\multicolumn{2}{c}{B0 Inc10}   & 
			\multicolumn{2}{c}{B50 Inc10}	& 
			\\  
			& 
			{$\bar{\mathcal{A}}$} & ${\mathcal{A}_B}$  
			& {$\bar{\mathcal{A}}$} & ${\mathcal{A}_B}$
			& {$\bar{\mathcal{A}}$} & ${\mathcal{A}_B}$ 
			&  {$\bar{\mathcal{A}}$} & ${\mathcal{A}_B}$  
			& {$\bar{\mathcal{A}}$} & ${\mathcal{A}_B}$
			& {$\bar{\mathcal{A}}$} & ${\mathcal{A}_B}$ 
			\\
			\midrule
			Finetune  &  3.16 & 0.96 & 1.72 & 1.05 & 7.84 & 4.44& 5.30 & 2.46& 3.14 & 1.10 & 1.54 & 1.13\\
			CoOp~\cite{zhou2022learning} & 14.54 & 7.14 & 13.05 & 7.77 & 47.00 & 24.24 & 41.23 & 24.12& 36.46 & 21.65& 37.40 & 20.87\\
			SimpleCIL~\cite{zhou2023revisiting} &59.24 & 48.09 & 53.05 & 48.09 & 84.15 & 76.63& 80.20 & 76.63& 92.04 & 86.85 & 88.96 & 86.85\\
			ZS-CLIP~\cite{radford2021learning} &26.66 & 17.22 & 21.70 & 17.22& 81.81 & 71.38& 76.49 & 71.38& 82.60 & 76.37& 78.32 & 76.37\\
			L2P~\cite{wang2022learning}  &47.19 & 28.29 &44.07&32.13& 82.74 & 73.03& 81.14 & 73.61& 76.63 & 61.82& 76.37 & 65.64 \\
			DualPrompt~\cite{wang2022dualprompt}  & 44.30& 25.83 &46.07&33.57 & 81.63 & 72.44& 80.12 & 72.57& 76.26 & 62.94& 76.88 & 67.55 \\
			CODA-Prompt~\cite{smith2023coda}  & 45.98 & 27.69 & 45.14 & 32.28& 82.43 & 73.43& 78.69 & 71.58& 80.21 & 66.47& 75.06 & 64.19 \\
			RAPF~\cite{huang2024class}   &  50.38  & 23.61 &  40.47 &  25.44 & 86.14 & 78.04 & 82.17 &  77.93  & 82.89 & 62.85 &  75.87 & 63.19\\\midrule
   		
			\name   & \bf69.94  &  \bf59.67 &  \bf65.34 &  \bf60.47 &  \bf86.41 & \bf79.48 & \bf83.68 &  \bf80.12 & \bf94.45 & \bf90.45 & \bf92.30 &  \bf90.91\\
		\end{tabular}
	}	
	\resizebox{1\textwidth}{!}{%
		\begin{tabular}{@{}lccccccccccccccc}
			\toprule
			\multicolumn{1}{c}{\multirow{3}{*}{Method}}
			& 
			\multicolumn{4}{c}{ImageNet-R }   & 
			\multicolumn{4}{c}{CUB }	&	\multicolumn{4}{c}{UCF }   
			\\ 
			& 
			\multicolumn{2}{c}{B0 Inc20}   & 
			\multicolumn{2}{c}{B100 Inc20}	&	\multicolumn{2}{c}{B0 Inc20}   & 
			\multicolumn{2}{c}{B100 Inc20}	& 
			\multicolumn{2}{c}{B0 Inc10}   & 
			\multicolumn{2}{c}{B50 Inc10}	& 
			\\  
			& 
			{$\bar{\mathcal{A}}$} & ${\mathcal{A}_B}$  
			& {$\bar{\mathcal{A}}$} & ${\mathcal{A}_B}$
			& {$\bar{\mathcal{A}}$} & ${\mathcal{A}_B}$ 
			&  {$\bar{\mathcal{A}}$} & ${\mathcal{A}_B}$  
			& {$\bar{\mathcal{A}}$} & ${\mathcal{A}_B}$
			& {$\bar{\mathcal{A}}$} & ${\mathcal{A}_B}$ 
			\\
			\midrule
			Finetune  & 1.37 & 0.43& 1.01 & 0.88& 2.06 & 0.64& 0.56 & 0.47& 4.51 & 1.59& 1.21 & 0.80\\
			CoOp~\cite{zhou2022learning} &60.73 & 37.52& 54.20 & 39.77& 27.61 & 8.57& 24.03 & 10.14& 47.85 & 33.46& 42.02 & 24.74\\
			SimpleCIL~\cite{zhou2023revisiting} & 81.06 & 74.48& 76.84 & 74.48& 83.81 & 77.52& 79.75 & 77.52& 90.44 & 85.68& 88.12 & 85.68\\
			ZS-CLIP~\cite{radford2021learning} &83.37 & 77.17& 79.57 & 77.17 & 74.38 & 63.06& 67.96 & 63.06& 75.50 & 67.64& 71.44 & 67.64\\
			L2P~\cite{wang2022learning}  &75.97 & 66.52 & 72.82 & 66.77&   70.87&57.93 & 75.64 &66.12 & 86.34 & 76.43& 83.95 & 76.62 \\
			DualPrompt~\cite{wang2022dualprompt}  &76.21 & 66.65 & 73.22 & 67.58&69.89 &57.46 & 74.40 &64.84 & 85.21 & 75.82& 84.31 & 76.35 \\
			CODA-Prompt~\cite{smith2023coda}  & 77.69 & 68.95 & 73.71 & 68.05& 73.12&62.98 &73.95&62.21 & 87.76 & 80.14&83.04 & 75.03 \\
			RAPF~\cite{huang2024class}  & 81.26  & 70.48 & 76.10 & 70.23 &  79.09 & 62.77& 72.82 & 62.93 & 92.28 & 80.33&90.31 & 81.55\\\midrule
			 \name    & \bf85.39 & \bf79.73  & \bf81.72 & \bf79.83 & \bf87.03 & \bf80.75& \bf83.13 & \bf80.87 & \bf93.22 & \bf88.08& \bf92.01 & \bf88.23\\
			
		\end{tabular}
	}
	
	\resizebox{1\textwidth}{!}{%
		\begin{tabular}{@{}lccccccccccccccc}
			\toprule
			\multicolumn{1}{c}{\multirow{3}{*}{Method}}
			&
			\multicolumn{4}{c}{SUN }   & 
			\multicolumn{4}{c}{Food }	&	\multicolumn{4}{c}{ObjectNet }   
			\\ 
			& 
			\multicolumn{2}{c}{B0 Inc30}   & 
			\multicolumn{2}{c}{B150 Inc30}	&		\multicolumn{2}{c}{B0 Inc10}   & 
			\multicolumn{2}{c}{B50 Inc10}	& 
			\multicolumn{2}{c}{B0 Inc20}   & 
			\multicolumn{2}{c}{B100 Inc20}	& 
			\\  
			& 
			{$\bar{\mathcal{A}}$} & ${\mathcal{A}_B}$  
			& {$\bar{\mathcal{A}}$} & ${\mathcal{A}_B}$
			& {$\bar{\mathcal{A}}$} & ${\mathcal{A}_B}$ 
			&  {$\bar{\mathcal{A}}$} & ${\mathcal{A}_B}$  
			& {$\bar{\mathcal{A}}$} & ${\mathcal{A}_B}$
			& {$\bar{\mathcal{A}}$} & ${\mathcal{A}_B}$ 
			\\
			\midrule
			Finetune  &4.51 & 1.59& 0.78 & 0.72& 3.49 & 1.71& 2.14 & 1.52& 1.34 & 0.47 & 0.69 & 0.54\\
			CoOp~\cite{zhou2022learning} &45.93 & 23.11 & 39.33 & 24.89& 36.01 & 14.18& 33.13 & 18.67& 21.24 & 6.29& 16.21 & 6.82\\
			SimpleCIL~\cite{zhou2023revisiting} & 82.13 & 75.58& 78.62 & 75.58& 87.89 & 81.65& 84.73 & 81.65& 52.06 & 40.13& 45.11 & 40.13\\
			ZS-CLIP~\cite{radford2021learning} &79.42 & 72.11& 74.95 & 72.11& 87.86 & 81.92& 84.75 & 81.92& 38.43 & 26.43 & 31.12 & 26.43\\	
			L2P~\cite{wang2022learning} &82.82 & 74.54 & 79.57 & 73.10 & 85.66 & 77.33& 80.42 & 73.13& 51.40 & 39.39& 48.91 & 42.83 \\
			DualPrompt~\cite{wang2022dualprompt} & 82.46 & 74.40 & 79.37 & 73.02& 84.92 &77.29& 80.00 & 72.75& 52.62 & 40.72& 49.08 & 42.92 \\
			CODA-Prompt~\cite{smith2023coda} &   83.34 & 75.71 & 80.38 & 74.17& 86.18 & 78.78& 80.98 & 74.13& 46.49 & 34.13& 40.57 & 34.13 \\
			RAPF~\cite{huang2024class}  & 82.13 & 72.47 & 78.04 & 73.10 & 88.57 & 81.15&\ 85.53 & 81.17&  48.67 & 27.43 & 39.28 &  28.73 \\\midrule
			 \name  & \bf84.89 & \bf78.24 & \bf82.34 & \bf79.25 & \bf89.04 & \bf83.05&\ \bf86.28 & \bf83.72&  \bf59.47  & \bf47.16  & \bf52.39 &  \bf47.31 \\
			\bottomrule
		\end{tabular}
  
	}

 	\caption{Average and last performance comparison of different methods.  
		The best performance is shown in bold.  All methods are initialized with the same pre-trained CLIP without exemplars for a fair comparison.}\label{tab:benchmark}

\end{table*}

%% file: Figs/curv.tex
\begin{figure*}[t]
    
\centering
\includegraphics[width=\linewidth]{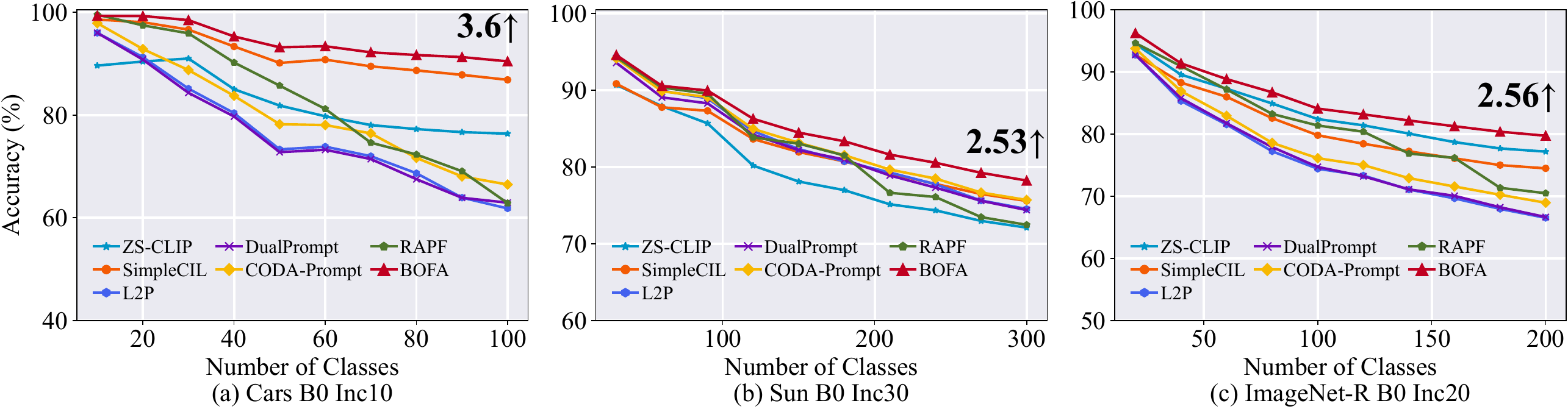}
    \captionsetup{skip=4pt}

    \caption{Incremental performance of different methods. Accuracy is reported at each incremental stage. \name~consistently outperforms all baselines, with the final gap to the strongest competitor noted at the end of each curve. }
    \label{fig:curve}
    \vspace{-6pt}
\end{figure*}

%% file: Figs/fig1.tex
\begin{figure}[t]
  \centering
  \includegraphics[width=\linewidth]{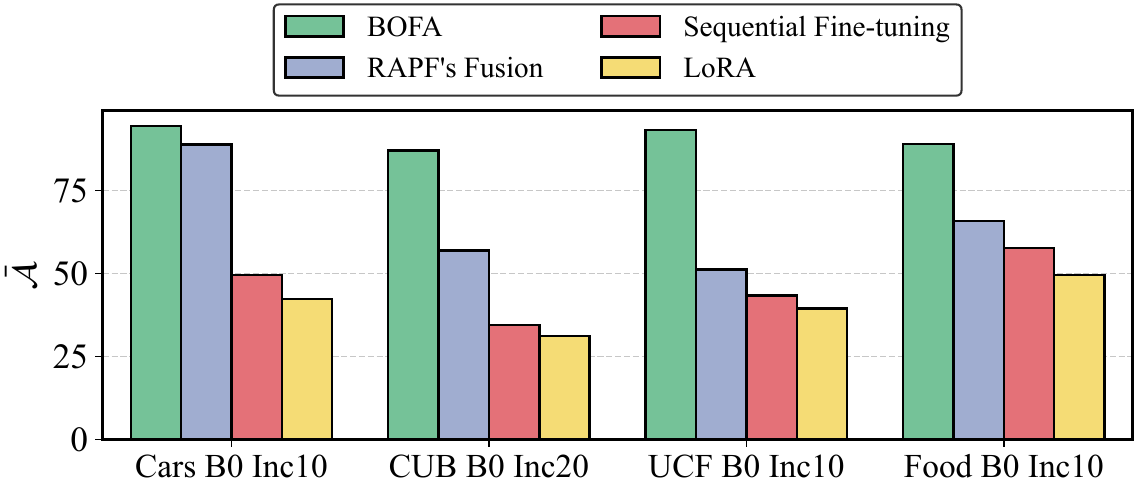}
    \captionsetup{skip=6pt}
  
  \caption{ $\bar{\mathcal{A}}$ comparison on four datasets for various ablation variants of \name. }

  \label{fig:ablation}

\end{figure}

%% file: Tabs/tab3.tex
\begin{table}[t]

	\centering
	\resizebox{\linewidth}{!}{

\begin{tabular}{@{}lcccccccc@{}}
\toprule
\multirow{3}{*}{\begin{tabular}[c]{@{}l@{}}Method \end{tabular}} &
  \multicolumn{4}{c}{ImageNet-R} &
  \multicolumn{4}{c}{SUN} \\ \cmidrule(l){2-9} 
 &
  \multicolumn{2}{c}{B0 Inc20} &
  \multicolumn{2}{c}{B100 Inc20} &
  \multicolumn{2}{c}{B0 Inc30} &
  \multicolumn{2}{c}{B150 Inc30} \\
 &
  $\bar{\mathcal{A}}$ &
  ${\mathcal{A}_B}$ &
  $\bar{\mathcal{A}}$ &
  ${\mathcal{A}_B}$ &
  $\bar{\mathcal{A}}$ &
  ${\mathcal{A}_B}$ &
  $\bar{\mathcal{A}}$ &
  ${\mathcal{A}_B}$ \\ \midrule
Textual &
  84.18 &
  78.53 &
  81.23 &
  79.22 &
  82.19 &
  74.82 &
  70.65 &
  77.07 \\
Visual &
  82.96 &
  75.68 &
  80.15 &
  76.98 &
  83.97 &
  76.38 &
  81.26 &
  77.40 \\ \midrule
 \textbf{\name} &
  \textbf{85.39} &
  \textbf{79.73} &
  \textbf{81.72} &
  \textbf{79.83} &
  \textbf{84.89} &
  \textbf{78.24} &
  \textbf{82.34} &
  \textbf{79.25} \\ \bottomrule
\end{tabular}
}
\captionsetup{skip=6pt}
	\caption{ Average and last performance comparison when using
 different prototype.} \label{tab:proto}

\end{table}

%% file: Figs/feat.tex
\begin{figure}[t]
  \centering
  \includegraphics[width=\linewidth]{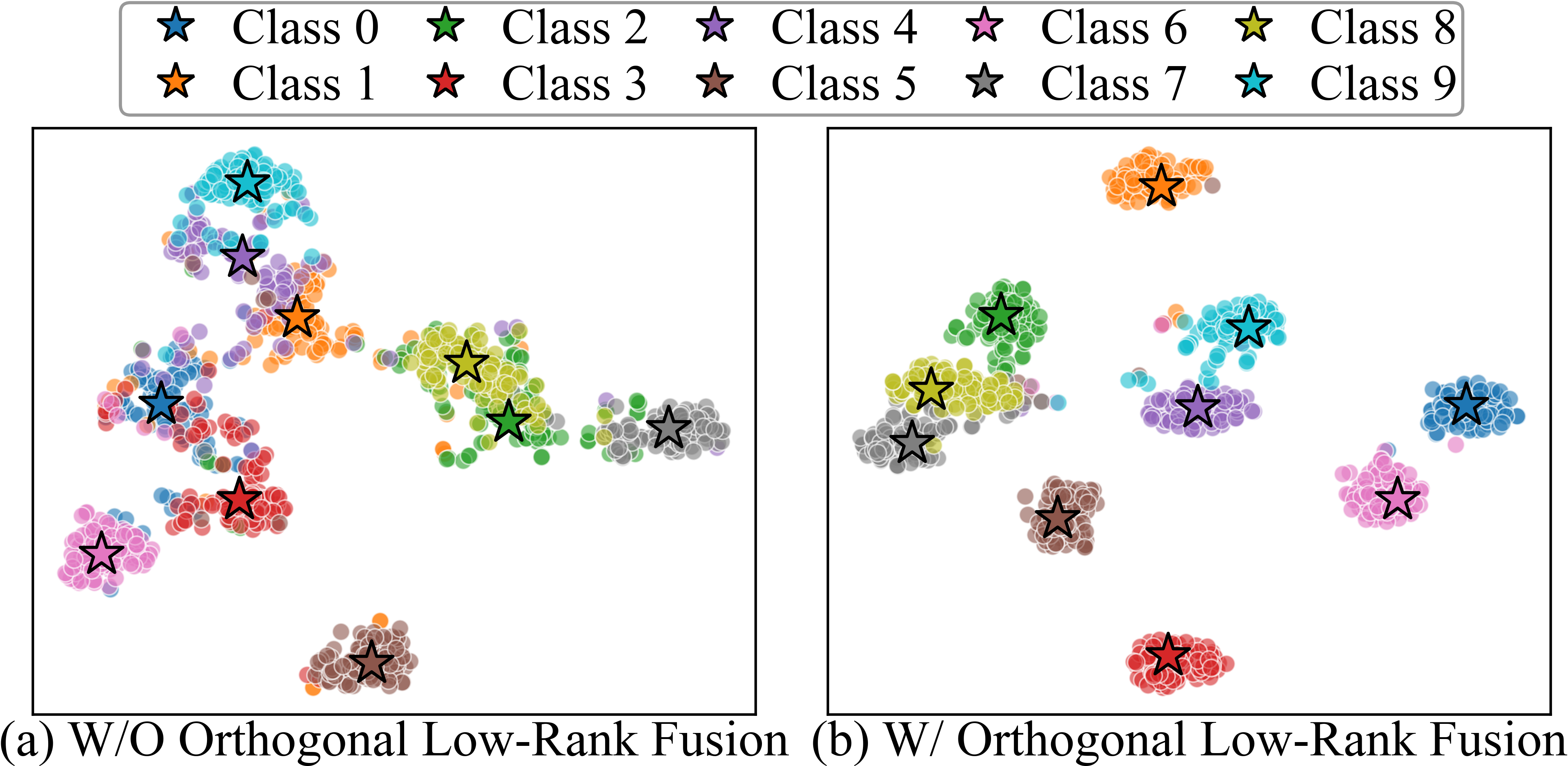}

  \captionsetup{skip=6pt}
  \caption{T-SNE visualization of features and class prototypes on CIFAR100 B0 Inc5. We show the feature distributions of old classes (0–4) and new classes (5–9) with (left) and without (right) applying Orthogonal Low-Rank Fusion. }

  \label{fig:tsne}
\end{figure}

%% file: _con.tex
\section{Conclusion}
In this work, we presented BOFA, an effective framework for exemplar-free CIL based on
CLIP. BOFA fine-tunes only CLIP's cross-modal bridge-layer, using a novel Orthogonal Low-Rank Fusion strategy to mitigate catastrophic forgetting without introducing extra parameters beyond the bridge-layer itself. Furthermore, it employs cross-modal hybrid prototypes to enhance classification by robustly integrating visual and textual cues. Our results show that this integrated approach effectively preserves knowledge while achieving SOTA discriminative performance.